# Inflatable Kirigami Crawlers

Burcu Seyidoğlu, Aida Parvaresh, Bahman Taherkhani, Ahmad Rafsanjani*

SDU Soft Robotics, Biorobotics Section, The Maersk McKinney Moller Institute, University of Southern Denmark, Odense 5230, Denmark

* Corresponding author ahra@sdu.dk

**ABSTRACT**

Kirigami offers unique opportunities for guided morphing by leveraging the geometry of the cuts. This work presents inflatable kirigami crawlers created by introducing cut patterns into heat-sealable textiles to achieve locomotion upon cyclic pneumatic actuation. Inflating traditional air pouches results in symmetric bulging and contraction. In inflated kirigami actuators, the accumulated compressive forces uniformly break the symmetry, enhance contraction compared to simple air pouches by two folds, and trigger local rotation of the sealed edges that overlap and self-assemble into an architected surface with emerging scale-like features. As a result, the inflatable kirigami actuators exhibit a uniform, controlled contraction with asymmetric localized out-of-plane deformations. This process allows us to harness the geometric and material nonlinearities to imbue inflatable textile-based kirigami actuators with predictable locomotive functionalities. We thoroughly characterized the programmed deformations of these actuators and their impact on friction. We found that the kirigami actuators exhibit directional anisotropic friction properties when inflated, having higher friction coefficients against the direction of the movement, enabling them to move across surfaces with varying roughness. We further enhanced the functionality of inflatable kirigami actuators by introducing multiple channels and segments to create functional soft robotic prototypes with versatile locomotion capabilities.

## 1. Introduction

Soft robots are constructed from flexible and lightweight materials that allows them to adapt to complex environments with enhanced safety and functionality [1]. Textiles, with their inherent flexibility, are widely used to create soft robots for diverse applications [2,3]. The McKibben artificial muscle, developed in 1960s by encapsulating pneumatic bladder in a rayon-made braid, is one of the earliest examples of employing textiles in soft actuators [4]. Since then, textiles have become a versatile component in a wide range of soft robotic applications. These include soft grippers [5-8], soft sensors [9-13], soft robotic hands [14-16], wearables [17-21], inflatables [22-26], and minimally invasive surgery devices [27-29].

Developing pneumatic soft actuators by fusing airtight fabric layers is an effective strategy for creating lightweight soft robots [30-34]. However, the thermoplastic coating that makes these textiles airtight also restricts their stretchability. Inflation of textile air pouches typically results in contraction and significant lateral bulging [24, 35]. Structuring air chambers into narrow, interconnected compartments can reduce bulging but also diminishes contraction. Moreover, the inextensibility of the textile often leads to nonuniform deformation and local buckling.

Kirigami metamaterials, created by introducing periodic cut patterns into thin sheets of material, enable the programming of deformation and mechanical properties [36]. These structures have recently gained attention for their use in soft robotic technologies, including adaptable grippers, locomotion, and wearables [36- 41]. Integrating a kirigami design approach into inflatable textile-based actuators can enable large, uniform deformation without excessive bulging or unpredictable buckling [42,43]. In addition to controlling global deformation, the coordinated local deformation of cuts in kirigami metamaterials allows for adjusting surface texture and frictional properties, which is beneficial for crawling robots [44-47]. Among crawling robots, many designs draw inspiration from origami, often relying on complex folding sequences and hinge mechanisms [48–50]. While both origami- and kirigami-based approaches utilize geometric patterning to enable actuation and mobility, kirigami offers the advantage of incorporating cuts into flat materials, thereby reducing mechanical constraints and simplifying fabrication.

Crawling locomotion in limbless creatures relies on two fundamental principles: body deformation through rhythmic muscular contractions and asymmetric friction achieved via skin features. These principles are exemplified in snakes' rectilinear and undulatory locomotion, where sequential contraction and relaxation of body segments, combined with the friction anisotropy of overlapping belly scales, allow for traversing diverse terrains [51,52]. Locomotive robots entirely made of textiles are rare, as most designs incorporate textiles alongside other materials or systems. However, there are notable examples where textiles play a pivotal role in crawling mechanisms. Recently, we developed an origami textile soft robot capable of rectilinear locomotion by integrating air pouches into the ventral side of a tubular origami structure featuring snakeskin-like pleats. These pleats unfold and engage with the ground upon periodic inflation, enabling propulsion in a straight line [53]. Mendoza *et al*. [54] developed an inflatable soft crawling robot made of textile-based actuators designed to inspect and diagnose power cables. This robot achieved bidirectional movement through two radial anchoring modules and a central longitudinal driving module, enabling it to navigate cables while performing fault diagnosis. Similarly, Kandhari *et al*. [55] introduced FabricWorm and MiniFabricWorm, textile-based soft robots inspired by earthworms capable of peristaltic locomotion for applications like search and rescue and pipe inspection. FabricWorm integrates textiles with 3D-printed parts and springs, while MiniFabricWorm achieves movement with no rigid parts. Guo *et al*. [56] developed soft textile robots with an internal PET bladder, combining textiles with nonlinear inflation for complex shape morphing inspired by mollusks and plants. They demonstrated continuous rolling, bidirectional crawling, and in-pipe crawling using a single pressure source by encoding sewing constraints onto highly elastic anisotropic fabric. Kennedy and Fontecchio [57] developed a textile-based robot actuated by bending shape memory alloy (SMA) wires, enabling locomotion through tumbling or rolling. Unlike conventional approaches that stitch SMA onto existing fabrics, they proposed a fully integrated woven hybrid SMA-textile actuator for functional soft robotic applications. These examples highlight the critical role of textiles in soft crawling robots, serving as the primary enabler of locomotion and forming the core of the robots' design and functionality, even when integrated with other materials.

This work presents a modular textile kirigami robot capable of multimodal locomotion observed in snakes. Our design builds upon the inflatable kirigami actuators which consist of a kirigami metamaterial with staggered linear cut patterns and an interconnected air pouch network embedded between two airtight textile layers. A distinctive feature of this class of actuators is that the cut pattern enables significant contraction through inflation-induced overlapping of sealed cut edges. We enhanced this design by connecting multiple modules and introducing double air channels within each module. Leveraging the overlapping morphology of inflated kirigami cuts, we achieved locomotion through friction modulation through cyclic inflation and deflation. By altering the actuation sequences of individual air channels, our system replicates various snake locomotion modalities, including rectilinear locomotion, serpentine motion (lateral undulation), and steering. We fabricated several variants of inflatable kirigami actuators, including single-channel and double-channel designs, and characterized their deformation experimentally and through finite element simulations. The functionality of our textile kirigami robot arises from its interaction with the environment and its ability to exhibit anisotropic deformation through structural reconfiguration. We evaluated its locomotion capabilities in both laboratory and outdoor environments under various scenarios, showcasing its versatility and adaptability.

## 2. Results and Discussion

*Inflatable kirigami actuators*. The building block of our crawling soft robot is an inflatable kirigami actuator, recently proposed by Chung *et al*. [42]. This actuator is fabricated by heat-pressing a layer of masking paper (Fig. 1A) between two TPU-coated nylon sheets, into which an array of staggered linear cuts is introduced using a laser cutting machine (see Figure S1 for detailed fabrication steps). Upon inflation of the kirigami actuator at $P = 100$ kPa, the air channels slightly bulge, and the edges of the cuts pass and overlap each other, resulting in a uniform contraction of $\varepsilon = -32\%$ (Fig. 1B). This contraction is almost twice as much as a similar air pouch without cuts that fails to perform a uniform contraction due to local buckling of inextensible pouches that gives rise to unpredictable post-buckling behavior (Fig. 1C), (see Video S1 for the comparison).

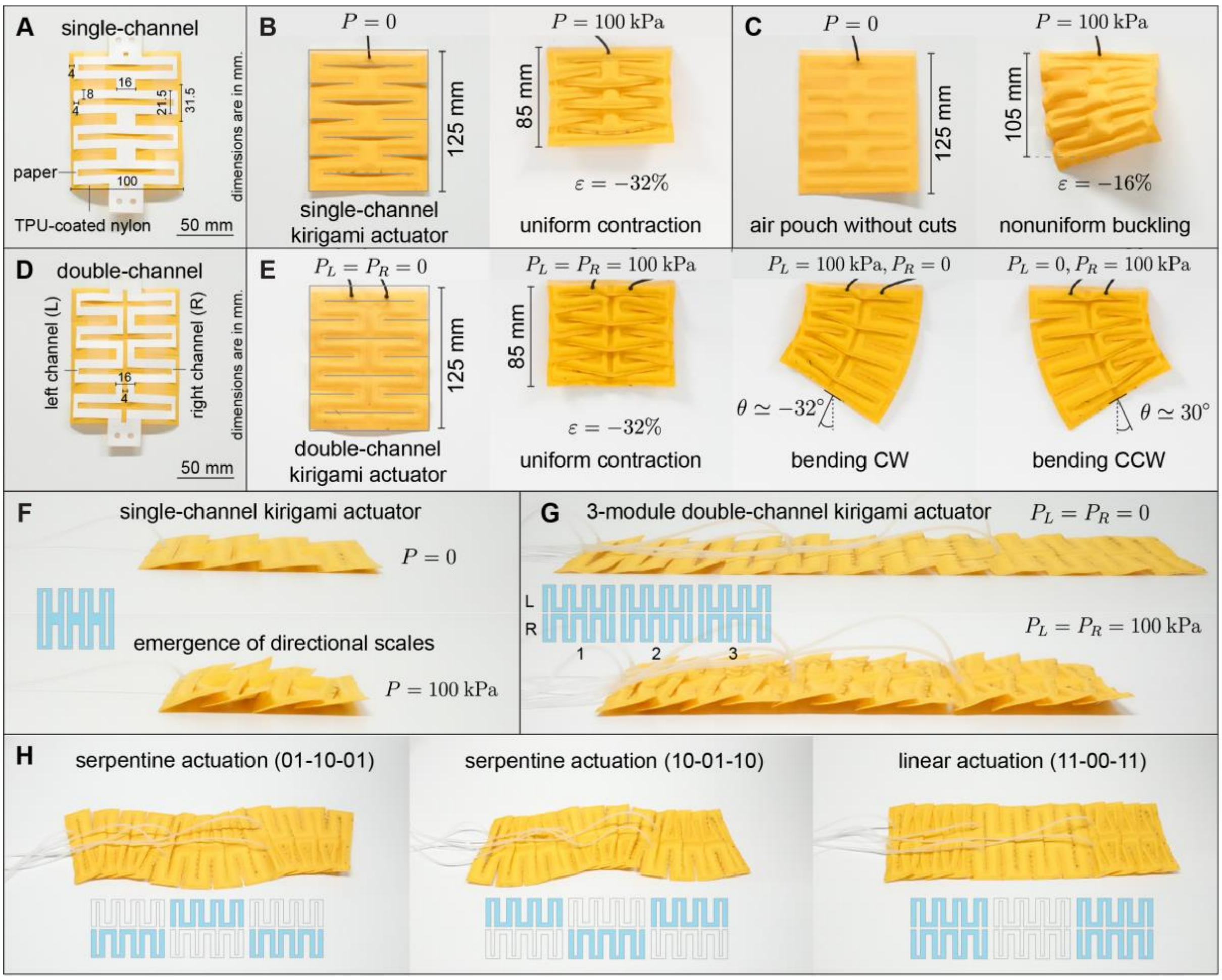

**Figure 1**. Inflatable kirigami actuators. (A) Single-channel kirigami actuator lamination with masking paper and TPU-coated nylon, (B) Uninflated (left) and inflated (right) single-channel kirigami actuator exhibiting uniform contraction, (C) Uninflated (left) and inflated (right) single-channel actuator (without cuts) experiencing nonuniform buckling under pressurization, (D) Double-channel kirigami actuator lamination with masking paper and TPU-coated nylon, (E) Uninflated double-channel kirigami actuator, uniform contraction when both channels are inflated, bending CW when left channel is inflated, and bending CCW when right channel is inflated (from left to right), (F) A module of single channel kirigami actuator in initial state (top) and when it is inflated (bottom) with emerging anisotropic scale-like features, (G) Three module double-channel kirigami actuator in uninflated state (top) and when all channels are inflated (bottom), (H) Examples of serpentine (left and middle) and rectilinear (right) gaits achieved by different sequences of actuation.

*Double-channel kirigami actuators*. For enhanced maneuverability and steering capability, we modified the design and integrated two parallel channels within the kirigami actuator, using the same materials and fabrication techniques (Fig. 1D). This actuator allows for independent inflation of two separate air channels, providing the ability to create differential actuation for steering and more complex motion profiles. Introducing an additional channel does not compromise the performance of the actuator, and the measured contraction is nearly identical to that of the single-channel kirigami actuator (Fig. 1E and Video S2). Differential inflation of this actuator allows producing repeatable bidirectional bending deformation.

**Emergence of scale-like features upon contraction**. The increased contraction of kirigami actuators is attributed to the rotation of pressurized channels and the contact-mediated self-assembly of the air pouches. When the rotation of these air pouches is coordinated, scale-like features emerge upon contraction, creating an asymmetry on the macroscopic surface texture. This asymmetry has the potential to generate directionally anisotropic friction, similar to the arrangement of ventral scales of a snake, which could be harnessed to enhance the locomotion capabilities of soft robots made of these actuators (Fig. 1F).

*Combining multiple modules*. We combined multiple double-channel kirigami actuators by arranging them in series, to create a crawling soft robot capable of a wide range of deformation modes. The three-module crawler features six independent air channels while maintaining the outer TPU layers as a single, continuous piece. This design simplifies fabrication without compromising functionality. Similar to single-channel actuators, simultaneous inflation of all channels causes the emergence of anisotropic scale-like features (Fig. 1G). By adjusting the inflation sequence across the six channels, this multimodal actuator can produce various locomotion gaits with a combination of flexural and axial contraction (see Video S3). For example, alternating inflation of air pouches within double-channel kirigami modules results in a serpentine motion, while symmetric inflation of both channels and applying different timing to each module can generate peristaltic linear locomotion (Fig. 1H).

*Contraction response*. To better understand the deformation behavior of kirigami actuators, we characterized their contraction profiles during cyclic inflation and deflation. The actuators were mounted vertically by fixing the top edge, and the pressure was ramped up to 100 kPa over 20 seconds, then ramped down to zero pressure over the same period, with a 5-second delay between each cycle. This process was repeated for 5 cycles. The recorded videos were analyzed using the Digital Image Correlation (DIC) library in MATLAB [58]. The contraction-pressure profiles of the single-channel and double-channel actuators are shown in Fig. 2A and Fig. 2B, respectively. A nonlinear relationship between the applied pressure and the contraction percentage was observed. At small pressures up to 40 kPa, the rate of contraction is higher during initial inflations, as the pouches within the kirigami structure can move freely in this pressure range. Beyond this pressure, the pouches bulge and self-assemble through local rotation and overlap while contacting each other, leading to a stiffening behavior that reduces the contraction rate. Both single-channel and double-channel kirigami actuators exhibited similar responses, although the hysteresis in the double-channel design was more pronounced. The lateral views of the kirigami actuators demonstrates the uniform contraction of kirigami actuators unlike non-kirigami air pouches that exhibit unpredictable deformation. While the characterization primarily focused on the inflation phase, we also examined the role of negative pressure (see Fig. S3), which is mainly used to efficiently vent the system and restore the actuator to its initial state. Its contribution to contraction is minimal, resulting in only a slight (4% for single-channel and 2% for double-channel) increase compared to when only positive pressure was applied.

*Bending response*. To characterize the steerability of the double-channel kirigami actuator, we measured the bending angle $\theta$ under dynamic differential inflation and deflation of two underlying channels as demonstrated in Fig. 2C. In each channel, the pressure was ramped up to 100 kPa over 20 seconds, then ramped down to zero pressure over the same period with a 25-seconds delay between each inflation/deflation cycle. There is a 20-second time shift between the actuation of two channels starting with the left channel. To measure the bending angle, we tracked the inclination of a line passing through three markers placed at the bottom edge of the actuator. This overlapping inflation was designed to observe how the bending range of each channel was influenced by the inflation of the other. We found that the bending angle fluctuated around $\theta \approx 32°$, with remained air in the channels contributing to residual bending angle of $\theta \approx 2°$ at the end of each cycle.

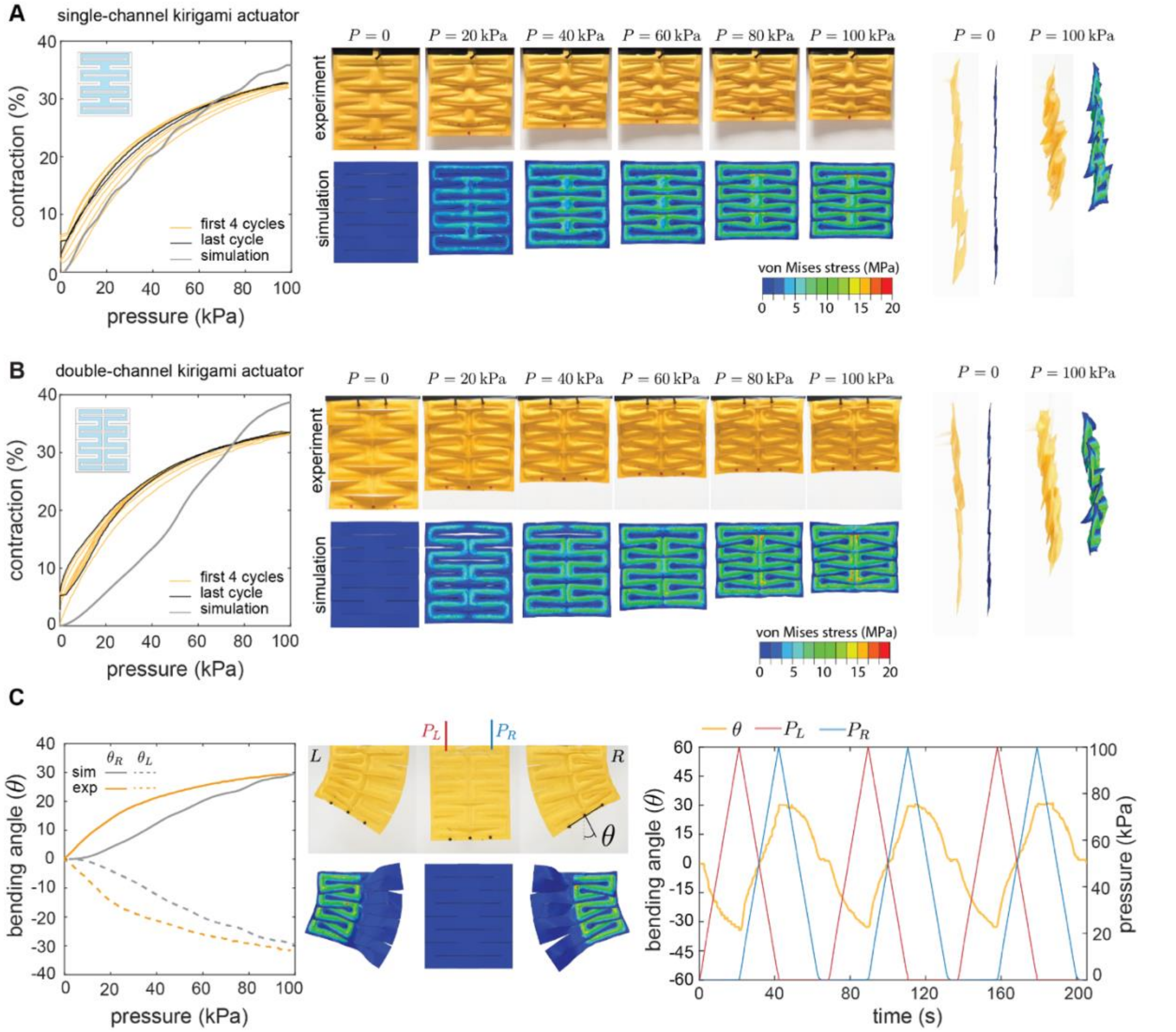

**Figure 2**. Mechanical response of the single module kirigami actuators. (A, B) Comparison of experimental and simulated pressure-contraction responses for single-channel (A) and double-channel (B) kirigami actuators (left). Corresponding snapshots of experiments and simulations at varying pressure levels (middle) and side views of the actuators in uninflated and inflated states (right). (C) Bending response of the double-channel kirigami actuator when each channel is inflated separately (left), corresponding snapshots from experiments and simulations (middle), and the bending angle of the double-channel kirigami actuator under cyclic, shifted ramp pressure inputs (right).

*Finite element simulations*. To gain deeper insight into the behavior of inflatable kirigami actuators, we simulated their mechanical response during inflation and deflation using the finite element (FE) method. Simulating kirigami actuators is particularly challenging due to their nonlinear deformation and self-contact behavior. However, we successfully modeled both single-channel and double-channel actuators, qualitatively reproducing trends observed in experiments.

The pressure-contraction curves from the simulations are overlaid with experimental results for both actuator types in Fig. 2A and Fig. 2B, showing fair agreement (see Video S4). The simulations indicate that the stiffening behavior at high pressures is primarily governed by the inextensibility of the textile material. Additionally, incorporating self-contact enabled accurate modeling of post-buckling behavior during contraction of overlapping scales. For double-channel actuators, bending behavior was simulated by applying differential inflation to the channels. As shown in Fig. 2C, the results accurately predict the bending trend despite some inevitable discrepancies, likely due to fabrication variability and simplified modeling assumptions, such as neglecting plastic behavior and the material's inherent anisotropy.

The simulations were based on a representative actuator geometry informed by our design constraints and fabrication limitations. Key design parameters including the width of the air channels, slit spacing, and the overall actuator aspect ratio, were selected to balance structural flexibility and directional deformation while ensuring reliable sealing and consistent fabrication. While not varied systematically in this study, these parameters are known to strongly influence contraction ratio, bending curvature, and frictional dynamics. Their selection was therefore guided by preliminary empirical testing to identify a configuration that would enable both effective linear and steering locomotion.

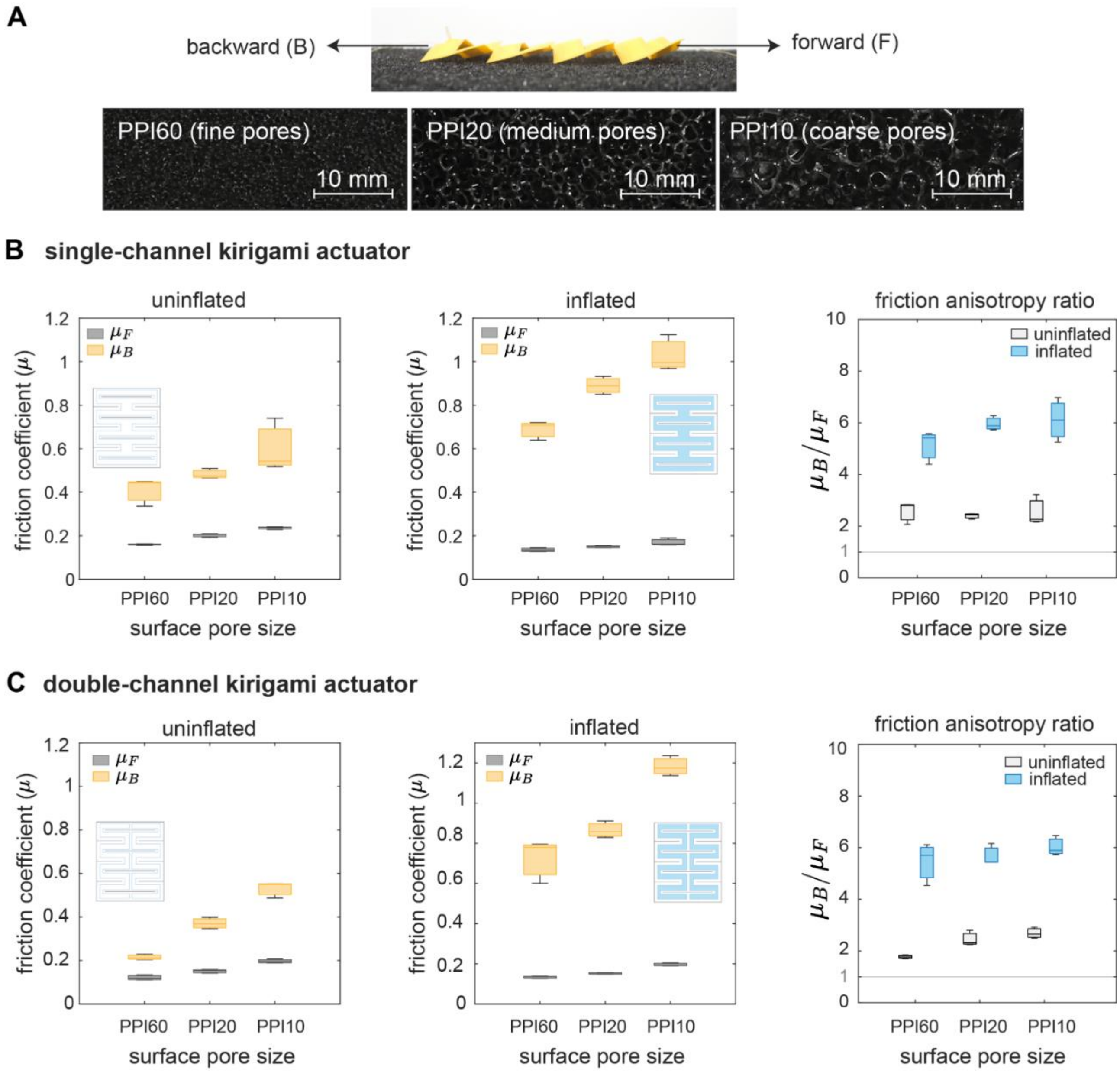

**Figure 3**. Friction response of the single module actuators. (A) Pulling directions for the friction tests (top) and three types of foam surfaces with different pore sizes used for the friction tests (bottom), (B) Measured friction coefficients in forward ($\mu_F$) and backward ($\mu_B$) directions of a single-channel kirigami actuator under uninflated (left) and inflated (middle) states (n=3), and the corresponding friction anisotropy ratios $\mu_B/\mu_F$, (C) Measured friction coefficients in forward ($\mu_F$) and backward ($\mu_B$) directions of a double-channel kirigami actuator under uninflated state (left) and when both channels are inflated (middle) (n=3), and the corresponding friction anisotropy ratios $\mu_B/\mu_F$.

*Friction response*. We characterized the frictional response of single-channel and double-channel inflatable kirigami actuators on surfaces with varying roughness. The resistive force was measured while pulling the actuators forward and backward relative to the orientation of scale-like features on polyurethane foams with large (PPI10), medium (PPI20), and fine (PPI60) open pores (Fig. 3A). Friction coefficients, $\mu_F$ and $\mu_B$ were obtained for the uninflated ($P = 0$) and inflated ($P = 100$ kPa) actuators and compared in Figs. 3B and 3C for the single-channel and double-channel designs, respectively. Additionally, we calculated the directional friction anisotropy ratio, $\mu_B/\mu_F$, for both designs across the three surfaces. The results indicate that, for all surfaces, $\mu_B > \mu_F$, confirming that the robot would crawl forward during cyclic actuation. Inflating the actuators had minimal effect on $\mu_F$ but nearly doubled $\mu_B$. Thus, inflation not only preserved frictional anisotropy but also enhanced it. As expected, the friction coefficients generally increased as the surface of the underlying foam substrates become coarser. Furthermore, the frictional responses of the single-channel and double-channel actuators were similar, with slight variations within the statistical range (see Note S1 for the friction analysis).

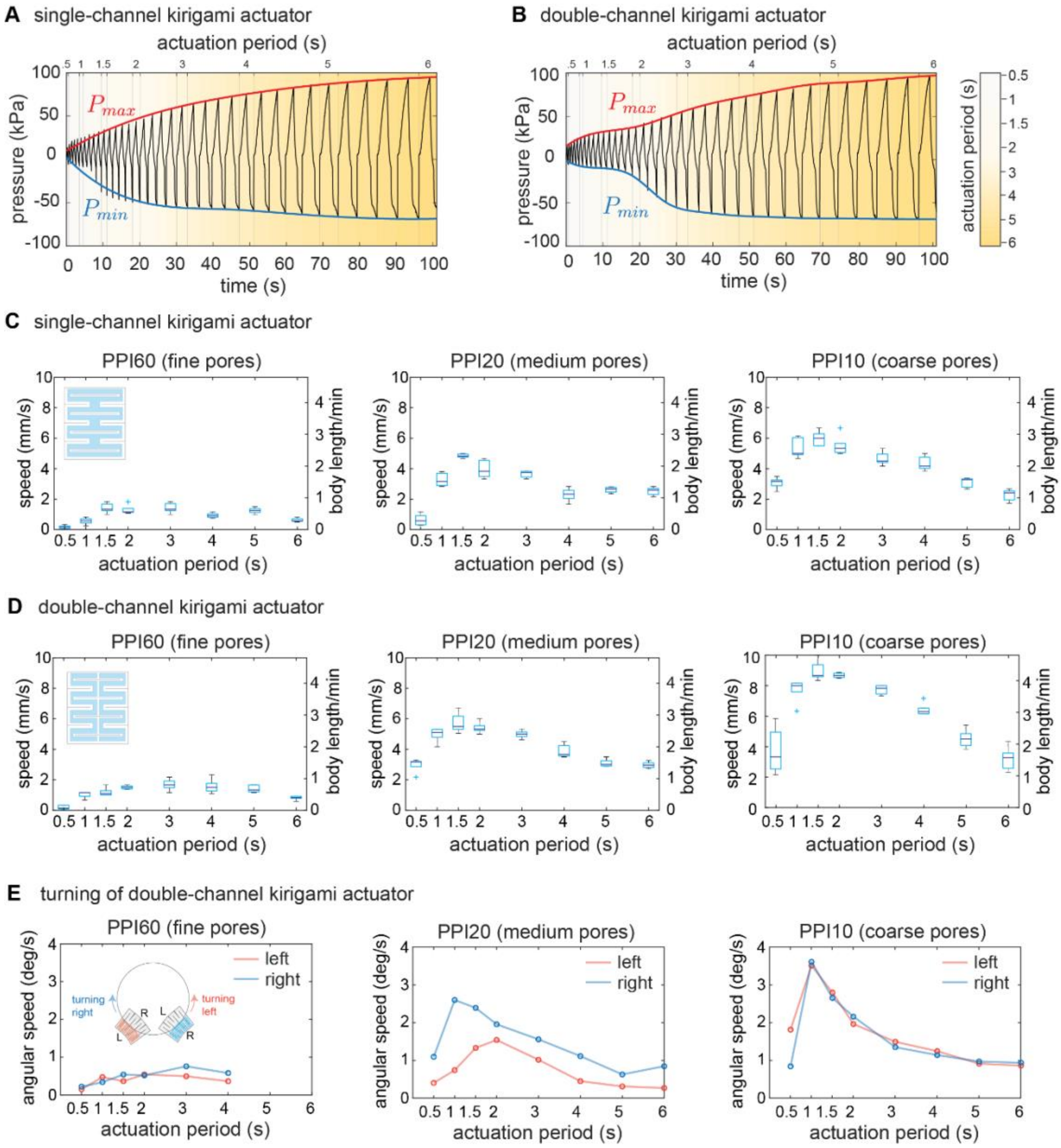

**Figure 4**. Characterization of the locomotion of kirigami actuators. (A, B) The variation of pressure levels of the single-channel (A) and double-channel (B) kirigami actuator with actuation period, (C, D) The linear locomotion speed of the single-channel (C) and double-channel (D) kirigami actuator under different actuation periods on different surfaces (n=5), (E) The average angular speed of the double-channel kirigami actuator across different actuation periods on various substrates.

*Locomotion of a single module*. We evaluated the linear locomotion of a single-module kirigami crawler in both single-channel and double-channel configurations, subjected to cyclic inflation and deflation at varying actuation periods on three foam substrates with different roughness (PPI60, PPI20, and PPI10). First, we varied the actuation period from $T = 0s$ to $T = 6s$ and tracked the internal pressure of the actuators for both designs, as shown in Fig. 4A and 4B, respectively. We observed that the pressure generally increases with actuation period, however with a faster rate at shorter period than longer periods. Particularly, around $T = 3s$ where $P \approx 50$ kPa, the changes in both positive and negative pressures become gradual. In pressure-contraction curves reported in Fig. 2A and 2B, we also observed that above this pressure level, the contraction rate declines. Based on these observations, operating around 50 kPa appears reasonable, as further increases in the pressure result in only minimal additional contraction.

Next, we considered eight actuation periods ($T = 0.5s, 1s, 1.5s, 2s, 3s, 4s, 5s, 6s$) and measured the linear speed on three foam substrates with different roughness levels during 1 minute of crawling (n=5). The results of these experiments for single-channel and double channel kirigami actuators are presented in Fig. 4C and Fig. 4D, respectively. We observed similar general trends across both designs and substrates. The robot crawls faster on rougher surfaces. This behavior indicates that enhanced friction exchange between the actuator and the surface leads to more efficient locomotion. Additionally, on each substrate, there is an optimal actuation period, which slightly shifts toward shorter periods as the surface roughness increases. This trend suggests a trade-off between the number of cycles and the stride length: shorter periods allow for more contraction/relaxation cycles within a given time, while longer periods enable larger contractions and potentially longer strides per cycle.

The single-channel kirigami actuator achieved maximum average speeds of $\bar{v} = 1.3$ mm/s, $\bar{v} = 4.8$ mm/s, and $\bar{v} = 6.0$ mm/s on PPI60 ($T = 3$s), PPI20 ($T = 1.5$s), and PPI10 ($T = 1.5$s) surfaces, respectively. Overall, the double-channel kirigami actuator demonstrated significantly better performance, with maximum average speeds of $\bar{v} = 1.6$ mm/s, $\bar{v} = 5.5$ mm/s, and $\bar{v} = 8.7$ mm/s on PPI60 ($T = 3$s), PPI20 ($T = 1.5$s), and PPI10 ($T = 1.5$s) surfaces. Given that the experimental conditions were identical for both designs, the higher locomotion speed of the double-channel actuator is likely due to more uniform contact facilitated with narrower central channels that reduced bulging (see Video S5 for comparison). The larger contact area could potentially strengthen engagement with surface asperities, reduce sliding, and enhance locomotion efficiency.

We also characterized the steering capability of the double-channel kirigami actuator across different actuation periods on various substrates (Fig. 4E). In these tests, we cyclically inflated and deflated only one channel of the double-channel actuator, repeating the procedure for both the left and right channels (see Video S6). On the PPI60 surface, we observed that the ground contact during single-channel inflation was insufficient to enable the actuator to complete a full rotation at longer actuation periods. On surfaces with coarser pores (PPI20 and PPI10), the steering behavior was more consistent, and we identified an optimal actuation period within the same range determined for linear locomotion ($T = 1 - 1.5$s). Some variation between the left and right channels was observed, which could be attributed to fabrication inaccuracies. We note that while only one channel was actuated in these tests to clearly isolate the steering mechanism, applying different pressure levels to the second channel could modulate the turning radius or adjust frictional interaction with the surface. However, in this study, we maintained the second channel unactuated to avoid introducing additional variables that might obscure the core relationship between asymmetric inflation and directional control.

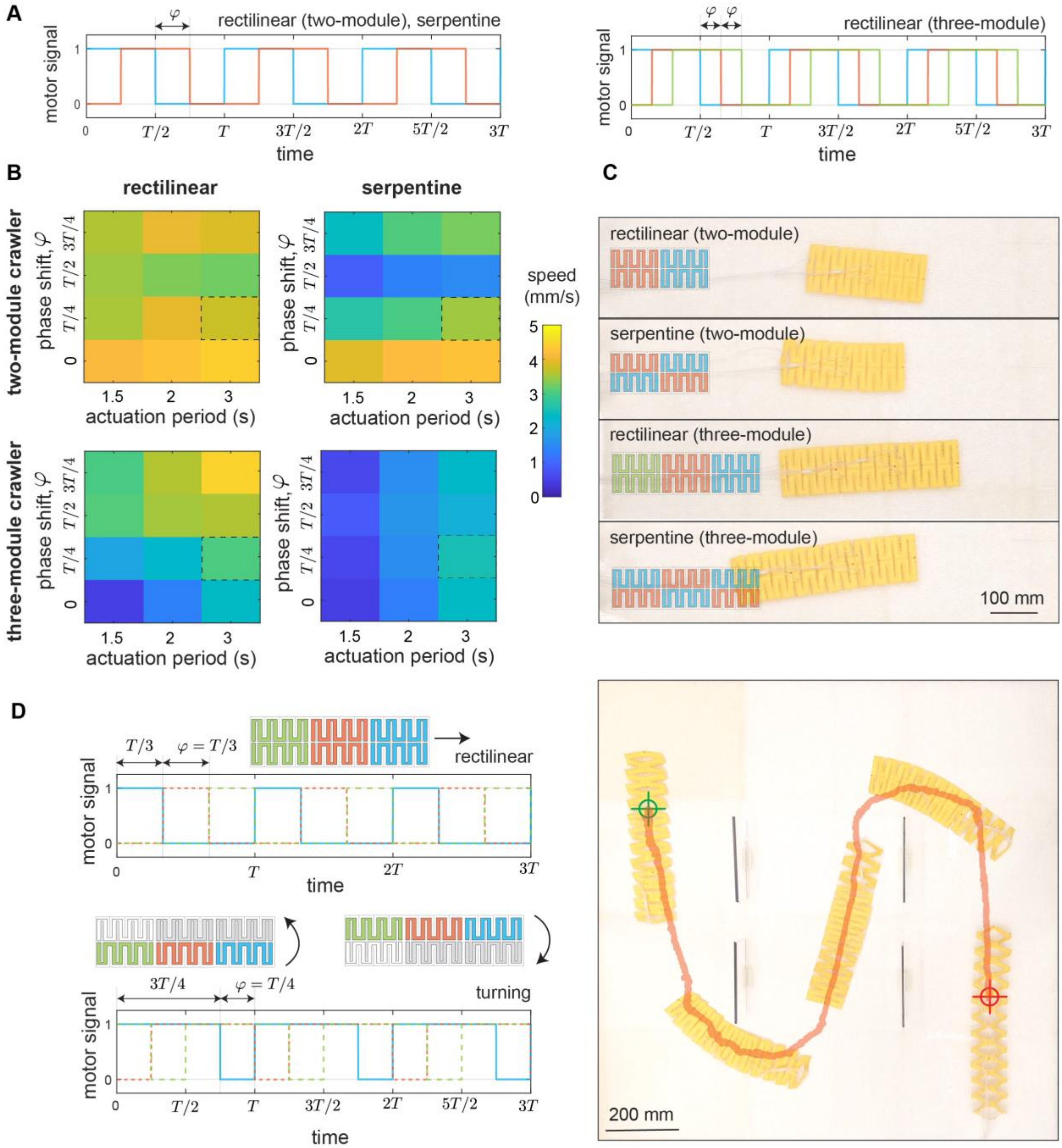

**Figure 5**. Characterization of the rectilinear and serpentine locomotion for multimodule kirigami robots. (A) Motor signals for serpentine locomotion of 2-module and 3-module kirigami crawlers, and rectilinear locomotion of the 2-module kirigami crawler (left), motor signals for rectilinear locomotion of 3-module kirigami crawler (right) with phase shift parameter, ϕ, (B)

the heatmaps showing the median speeds (n=5) of 2-module and 3-module kirigami crawlers for rectilinear and serpentine gaits with varying phase shifts of actuation, (C) the snapshots of the robots in their final position after 2 minutes with different gaits, (D) Motor signals for generating rectilinear and turning gaits to navigate an S-shaped path through obstacles (left), the snapshots of the completed path (in 19 minutes).

*Multi-module kirigami crawlers*. Connecting multiple kirigami crawlers in series and actuating them sequentially generates traveling waves characterized by an actuation period $T$ and a phase shift $\varphi$. We created two sets of crawlers, each consisting of two or three double-channel kirigami actuators, and evaluated their locomotion performance. We selected three actuation periods ($T = 1.5s, 2s, 3s$) corresponding to the optimal performance of single-module kirigami actuators, and considered four different phase shifts ($\varphi = 0, T/4, T/2, 3T/4$). Two locomotion modes, referred to as *rectilinear* and *serpentine* gaits, were examined by pairing different pressure inlets. For rectilinear locomotion, both channels within each module were inflated simultaneously, and a head-to-tail traveling wave was generated by introducing a phase shift between neighboring modules. For serpentine locomotion, the opposite channels were wired together across alternate modules to produce wavy motion through cyclic inflation. In this configuration, the phase shift corresponds to the delay between the activation of the two groups of interconnected channels. In these experiments, we considered actuation protocols with a half-period duty cycle as demonstrated in Fig. 5A (see Video S7).

For all four cases, the average speed was measured across considered actuation periods and phase shifts while crawling over a foam substrate with coarse pores (PPI10) for a duration of 2 minutes (n=5). The results are summarized in Fig. 5B and the snapshots of crawling locomotion for $T = 3s$ and $\varphi = T/4$ are shown in Fig. 5C. When $\varphi = 0$, both rectilinear and serpentine gaits, despite their different names, are identical as all the actuators inflate and deflate simultaneously. Overall, rectilinear locomotion was faster than the serpentine gait, and the two-module crawler outperformed the three-module configuration. Increasing the actuation period led to faster locomotion. This effect is more pronounced in the three-module crawler, as its greater weight necessitates stronger contractions to overcome friction. Additionally, its larger volume requires a longer inflation time, resulting in improved performance at higher actuation periods.

Introducing phase shifts had distinct effects on locomotion. The slowest performance was mostly observed when $\varphi = T/2$. For two-module crawler, $\varphi = T/4$ and $\varphi = 3T/4$, resulted in a very similar performance as over cyclic movement a similar wave emerges. In contrast, for the three-module robot, the phase shift pattern and the order in which actuators were activated had a more pronounced effect on overall locomotion. Generally, increasing the phase shift led to higher speeds, suggesting that inflating specific sections of the body in sequence helped mitigate the challenges posed by weight and friction, rather than contracting the entire body at once. This result revealed that the middle module has a key role in performance. For both $\varphi = T/4$ and $\varphi = 3T/4$, the first and the third modules followed the same actuation pattern, while the activation timing of the middle module varied, which appeared to influence traction, weight distribution and propulsion efficiency.

*Crawling through obstacles*. We demonstrated that the three-module crawler can navigate obstacles by following a complex path. Through experimentation, we optimized an actuation protocol for rectilinear gait, achieving greater efficiency with a duty cycle of $T/3$ and a phase shift of $\varphi = T/3$. In this configuration, modules inflate sequentially with no overlap in their inflation periods. For steering, we tested various duty cycles and found optimal performance at $3T/4$ with $\varphi = T/4$. Using these protocols, we manually controlled the robot to follow an S-shaped path around rigid walls while crawling over a foam substrate with coarse pores (PPI10) as demonstrated in Fig. 5D (see Video S8).

*Crawling across diverse terrains*. We demonstrated the ability of our three-module, double-channel kirigami crawling robot to traverse a variety of challenging surfaces. These tests included crawling on asphalt (Fig. 6A), maneuvering through a narrow channel (Fig. 6B), climbing inclines (Fig. 6C and 6D), navigating rough concrete with small gaps (Fig. 6E), and locomoting over a metal grating (Fig. 6F). These experiments highlight the remarkable adaptability of our inflatable kirigami crawlers. Despite their simple design, they effectively adjust to diverse terrain conditions, showcasing their potential for robust, versatile locomotion in unstructured environments (see Videos S9-S14 for locomotion on diverse terrains).

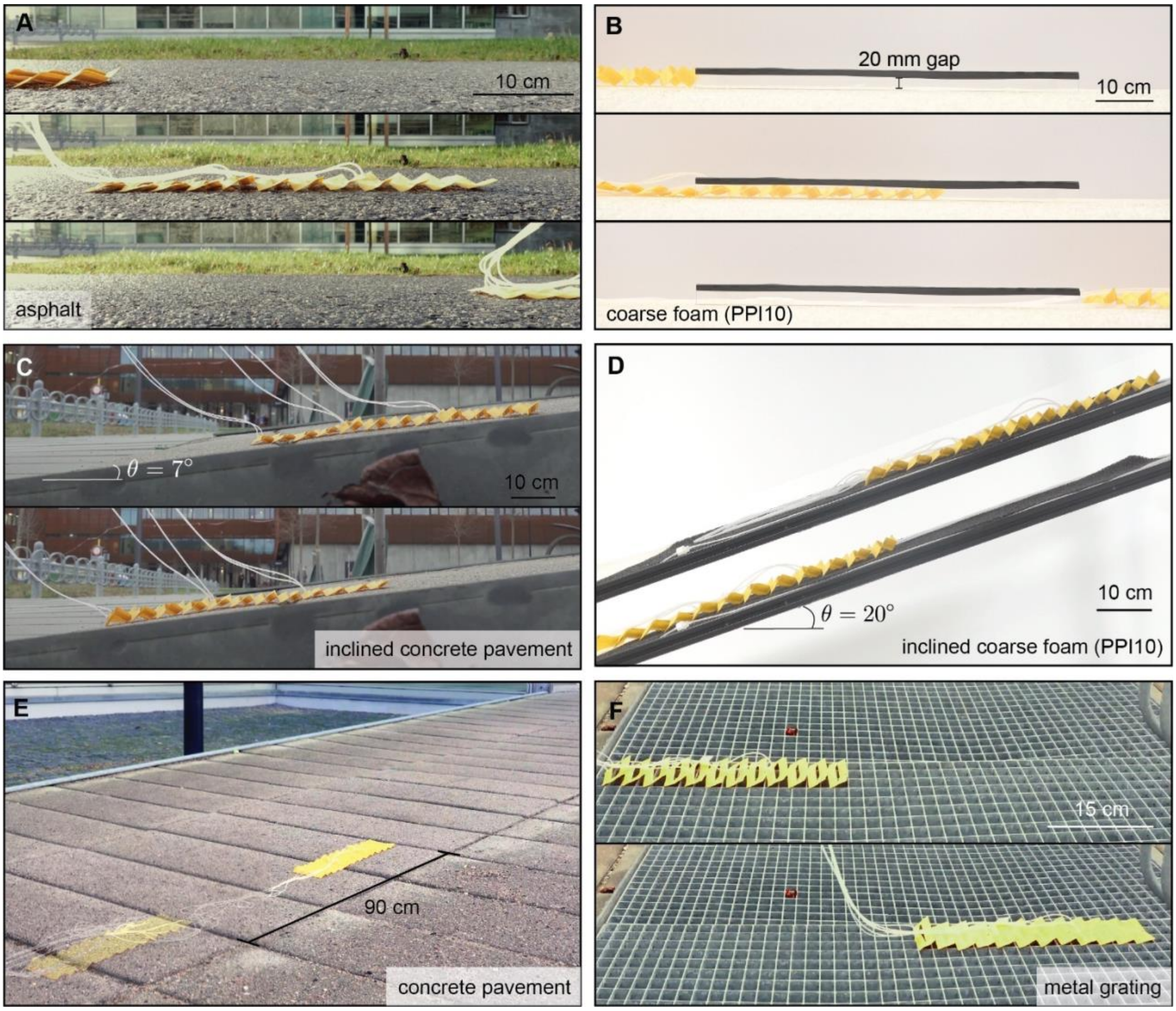

**Figure 6**. Locomotion of multimodule kirigami robot on diverse terrains. (A) Locomotion on asphalt, (B) passing through a narrow channel over a foam surface, climbing (C) a 7° inclined concrete pavement, and (D) a 20° inclined coarse foam substrate, crawling on (E) concrete pavements with gaps, and (F) over a metal grating.

## 3. Conclusion

In summary, this work purposefully exploits the shape-morphing behavior of inflatable kirigami metamaterial actuators to achieve multiple functions within a monolithic structure, enabling versatile locomotion capabilities. On one hand, the presence of cuts allows the kirigami actuator to function as an artificial muscle, capable of large-amplitude contractions at lower pressures compared to conventional air pouches. On the other hand, the kirigami design triggers nonlinear shape-morphing behavior, creating a uniform texture upon contraction and breaking symmetry to produce a directionally anisotropic friction response. By combining these two characteristics through cyclic actuation, we introduce a new class of crawling soft robots, entirely composed of airtight textile materials, capable of navigating various terrains with moderate surface roughness. We demonstrated the versatility of our proposed design by embedding multiple air channels within the kirigami actuators, which can be independently actuated to achieve more complex locomotion gaits, such as rectilinear locomotion, serpentine movement, and steering. Through a comprehensive set of experiments, we characterized the effects of single- versus double-channel designs, substrate surface roughness, actuation period, and the number of modules on the locomotion performance of the inflatable kirigami crawlers. Finally, we leveraged this knowledge to demonstrate the locomotion capabilities of the proposed kirigami crawlers across diverse scenarios.

Although the current setup with miniature air pumps and low-power solenoid valves requires tethering, the components are compact, lightweight, and energy-efficient—promising for future untethered designs. Untethered locomotion will depend on careful integration, particularly managing payload and weight distribution, which affect mobility and stability. As module count increases, coordinating multiple pneumatic channels becomes more complex. Our current 3-module crawler uses six solenoid valves for six air channels; scaling this linearly adds control overhead. Future designs could address this with pneumatic multiplexing and hierarchical control, reducing valve outputs while preserving actuation flexibility. These strategies are key to advancing adaptable kirigami crawlers.

## 4. Experimental Section

*Fabrication*. Kirigami actuators were fabricated using laser cutting and heat press lamination techniques. Each actuator was made from two layers of TPU-coated nylon (210 den, TPU-coated on one side, 275 g/m$^2$, heat-sealable), which were laser-cut with the kirigami pattern using a VLS/ULS 2.30 Universal Laser System. A standard office paper (Navigator Universal, 80 g/m$^2$) was also laser-cut to form air channels and precisely positioned between the TPU-coated nylon layers, with the coated sides facing inward for optimal adhesion. The assembly was secured with paperclips before being heat-pressed at 160°C for 10 seconds using a TC5 SMART SECABO machine. To prevent overheating, a cloth was placed between the heater plate and the material, while the tubing section remained outside the press. After the initial heat sealing, 3D resin-printed tubing was inserted to create inlets, followed by a second heat press using the same parameters. This process produced symmetric kirigami actuators without a preferred pop-up direction. To break symmetry and ensure uniform overlapping scales, the actuators were suspended with a 100 g mass, allowing the scales to form consistently when pressurized.

*Control System*. The crawlers were controlled using an Arduino UNO microcontroller, operating two miniature pneumatic diaphragm positive pressure and vacuum pumps (VN2708PM). Airflow was managed through six 3-way miniature pneumatic solenoid valves (Parker X-Valve, 8mm), with timing and sequence variations regulating the duration and direction of airflow during actuation (see Figures S2 and S3 for the schematics). Different gait patterns were achieved by interconnecting air channels in various configurations. All locomotion experiments were conducted using two diaphragm pumps (VN2708PM), which are responsible for the actuation of the inflatable kirigami crawlers.

*Pneumatic Characterization*. Characterizations of mechanical responses were performed using a microfluidic control system (Fluigent LineUp™), enabling independent measurement of contraction and bending without pump-related variations. Supplied with pressure via a compressor, the system allowed for precise regulation and control while also supporting multiple outlets. The devices were operated using Fluigent's SDK development kit integrated with MATLAB, ensuring accurate and customizable control.

*Friction measurements*. The robots were attached to a load cell (FUTEK LSB200, 5 lbs.) via a Kevlar thread and mounted on a motorized linear stage (Thorlabs LTS300). Each robot was pulled at a constant speed of 10 mm/s for 20 seconds in both forward and backward directions over three polyurethane foam surfaces with different pore size distributions: fine (PPI60), medium (PPI20), and coarse (PPI10). Force data was recorded using a National Instruments DAQ (NI USB-6008). A MATLAB script controlled the displacement of the linear stage and logged the data, ensuring accurate and consistent measurements.

*Finite element simulations*. The simulations were performed in the commercial nonlinear finite element software ABAQUS 2024. The actuators were modeled as two sheets, each with the same geometry and cut patterns as those used in the experiments. These sheets were discretized with linear shell elements (Element Type S3) with an average mesh size of 2 mm, and the bonded regions were connected using a Tie constraint. The material was assumed to be isotropic and elastic, characterized by a Young's modulus of $E = 230$ MPa and a Poisson's ratio of $\nu = 0.4$ . In all simulations, the top edge of the model was fixed, and a general contact model was applied to accurately capture realistic deformation profiles. The simulations were conducted using the Implicit Dynamic Solver. The axial simulation was divided into two steps. First, a small force couple was applied at the center of the middle slots to ensure uniform deformation and to slightly open the cuts. Then, a uniform pressure of $P = 100$ kPa was applied to the internal surfaces to induce contraction during inflation, leading to overlapping of the scales.

**Supporting Information**
Supporting information is accompanied this article.

**Acknowledgements**
This work was supported by the Villum Foundation through the Villum Young Investigator grant 37499. We thank Cao Danh Do for assistance with video recording.

**Conflict of Interest**
The authors declare no conflict of interest.

**Data Availability Statement**
The data that support the findings of this study are available from the corresponding author upon reasonable request.

Supporting Information

# Inflatable Kirigami Crawlers

Burcu Seyidoğlu, Aida Parvaresh, Bahman Taherkhani, Ahmad Rafsanjani*

SDU Soft Robotics, Biorobotics Section, The Maersk McKinney Moller Institute, University of Southern Denmark, Odense 5230, Denmark

* Corresponding author ahra@sdu.dk

This document contains:

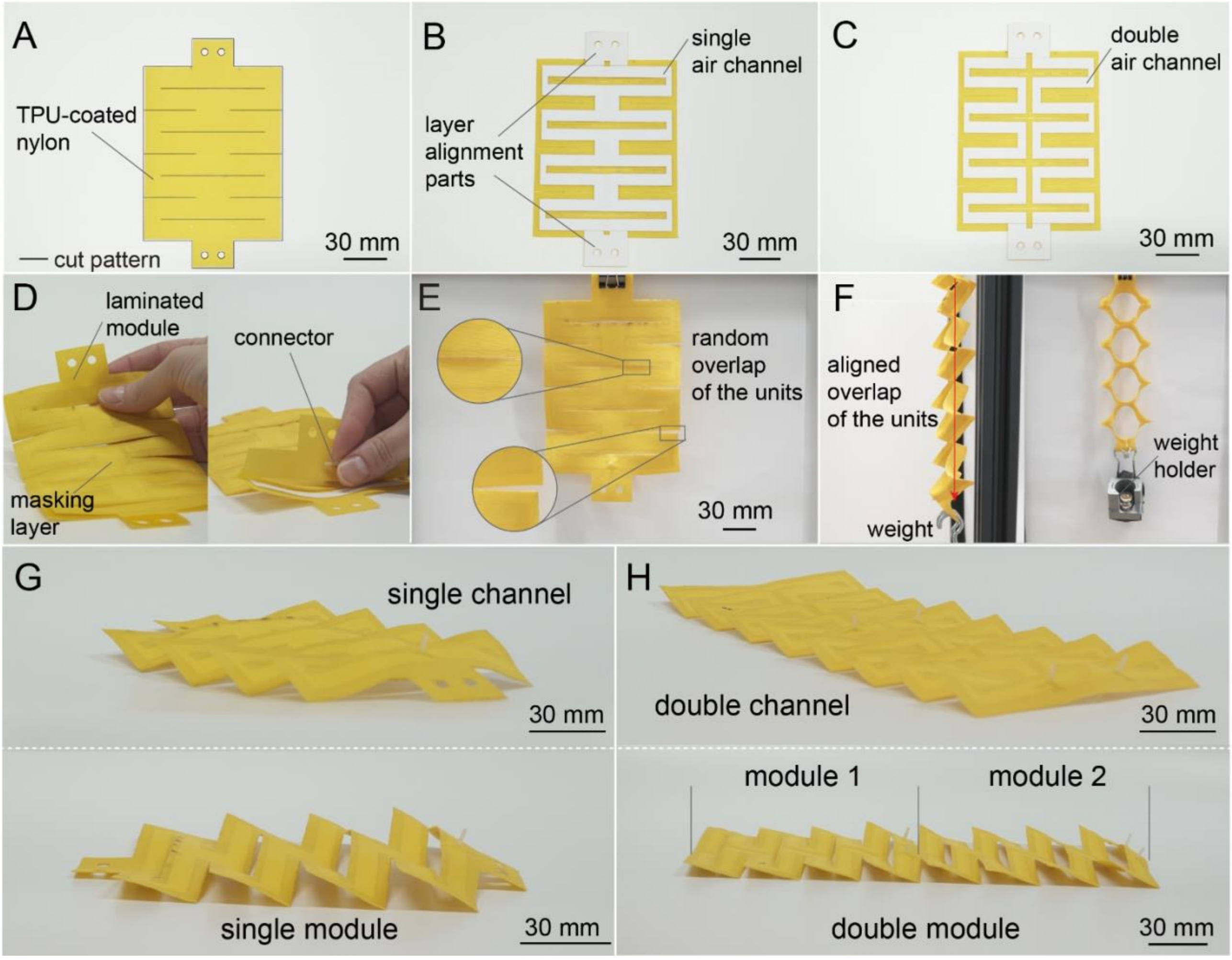

**Figure S1. Fabrication of inflatable kirigami crawlers**. (A) Two layers of TPU-coated nylon (210 den, TPU-coated on one side, 275 g/m$^2$, heat-sealable) are laser cut with the kirigami pattern, (B) standard office paper (Navigator Universal, 80 g/m$^2$) as masking layer for single-channel design, (C) for double-channel design, (D) insertion of tubing after initial heat sealing, (E) random alignment of the units after fabrication, (F) suspension with a 100 g mass to ensure uniform overlapping scales, (G) complete single-channel single module kirigami actuator, (H) complete double-channel 2-module kirigami crawler.

**Note S1. The variation of pressure levels with actuation period**

To further investigate the relationship between actuation period and internal pressure dynamics, which are originally shown in Figures 4A and 4B, we repeated the experiments under controlled pressure conditions using the Fluigent LineUp™ microfluidic pressure controller. These tests were conducted to decouple the actuator's mechanical response from the flow limitations of the diaphragm pump (VN2708PM) used in the locomotion experiments.

We tested both the single-channel and double-channel kirigami actuators under biphasic pressure input, where pressure cycled between 100 kPa and -80 kPa while the actuation period was swept from 0.5 s to 6 s. The selected pressure limits represent the maximum performance range of the microfluidic system and ensure a consistent, repeatable pressure input.

In these tests, we observed that both the single-channel and double-channel actuators exhibited incomplete pressurization at short actuation periods. At periods below 1 s, the actuators reached only ~80 kPa during inflation and ~–50 kPa during deflation, indicating that the system did not have sufficient time for full inflation/deflation. However, at longer periods (>1.5 s), the actuators achieved more balanced cycling between ~-70 and 100 kPa.

These results confirm that the flow rate limitations of the diaphragm pump influence the pressure dynamics observed in the locomotion experiments. Nevertheless, the trend of increasing pressure amplitude with longer actuation periods persists even with a precisely controlled pressure source. This suggests that while pump-induced flow constraints contribute to the behavior, the actuators' internal compliance and fluidic resistance also play a substantial role.

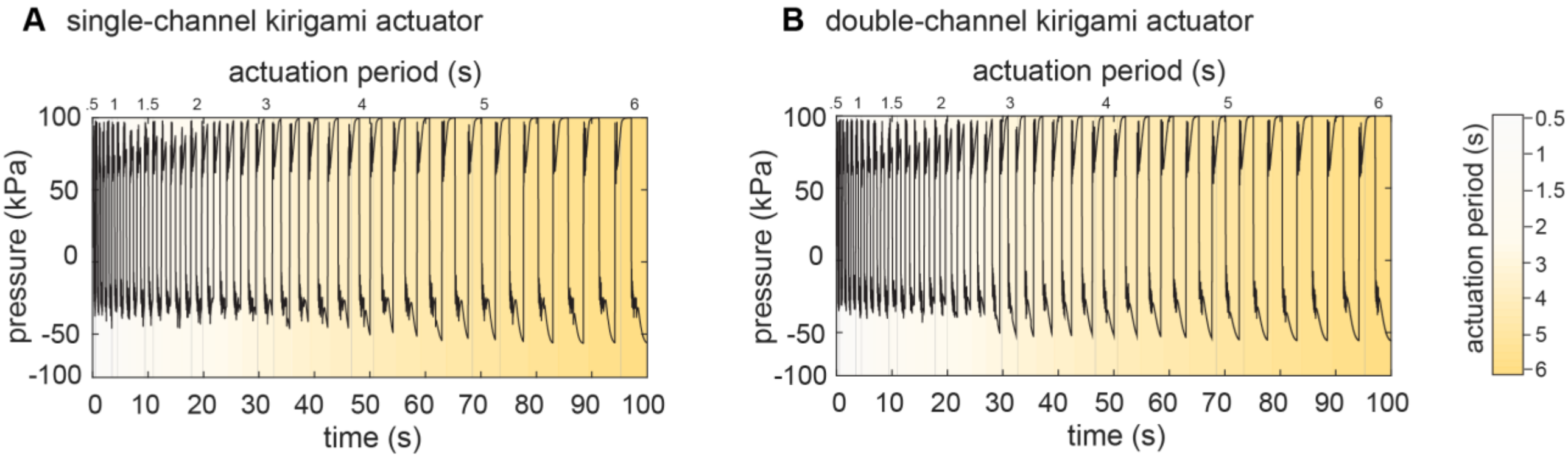

**Figure S2. The variation of pressure levels with actuation period**. The single-channel (A) and double-channel (B) kirigami actuators' pressure trackabilities under both positive and negative pressure, ranging from 100 kPa to -80 kPa, sweeping the actuation period from 0.5 s to 6 s.

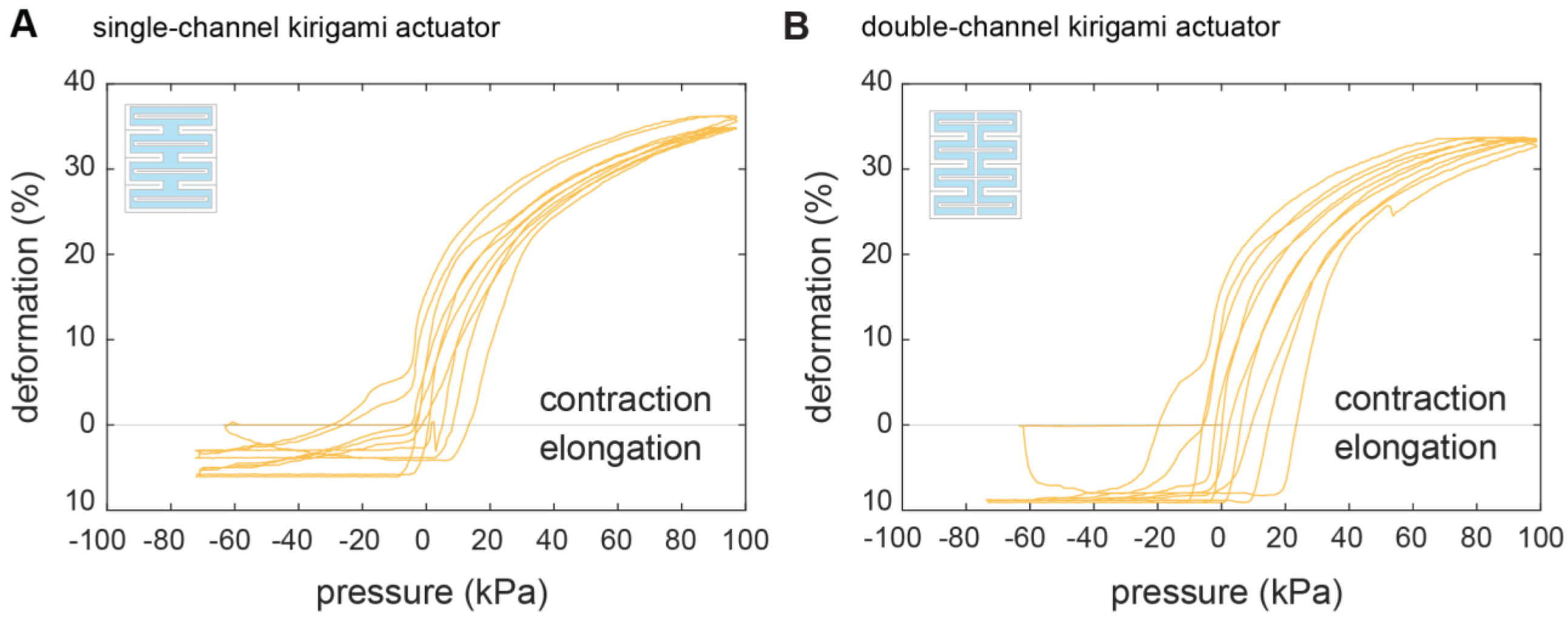

**Figure S3. Deformation of the actuators under both positive and negative pressure**. Pressure-contraction responses for single-channel (A) and double-channel (B) kirigami actuators under both positive and negative pressure.

**Note S2. Tensile strength and Blocked force of the kirigami actuators**

We performed blocked force measurements for both the initial (unstretched) and pre-extended (75 mm stretched from the initial length) configurations of the actuators while gradually increasing the internal pressure in 25 kPa steps from 0 to 100 kPa using a microfluidic control system. At each pressure level, we recorded the force output under a fixed-length (blocked) condition using a tensile testing machine. This setup allowed us to map the relationship between internal pressure and the corresponding blocked force at two distinct strain levels (0 and 0.6) corresponding to the initial length and the extended length (200 mm), respectively, with the initial actuator length being 125 mm.

The results revealed a nonlinear blocked force behavior, attributed to the out-of-plane deformations induced by the kirigami patterns and the fabric's compliant structure. As expected, increasing internal pressure resulted in higher blocked forces. Furthermore, higher strain levels also produced greater blocked forces, indicating that pre-stretching enhances the actuator's ability to exert force against a constraint.

Both single-channel and double-channel kirigami actuators exhibited similar trends in force response, demonstrating consistent mechanical behavior across designs. However, the double-channel actuators consistently generated slightly higher output forces across pressure levels, likely due to the presence of an additional pressurizable channel contributing to increased force generation.

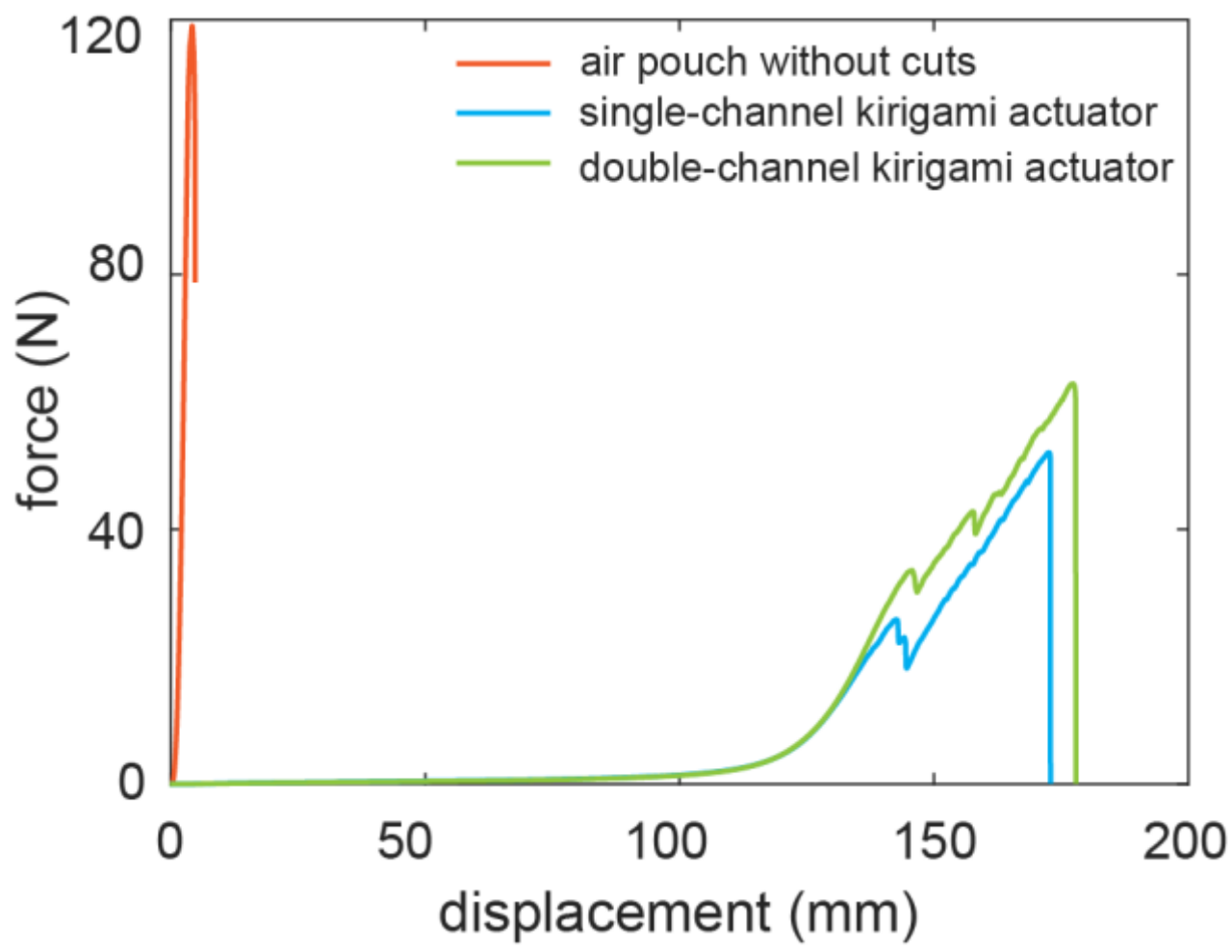

**Figure S4. Tensile test results of the air pouches without cuts and kirigami actuators**. The tensile test results or air pouch without cuts, single-channel kirigami actuator, and double-channel kirigami actuator. Tensile test results confirm that introducing kirigami cuts reduces the overall tensile strength compared to the uncut air pouch. The tests were terminated upon the onset of plastic deformation or minor tearing to capture the structural limits without causing complete rupture.

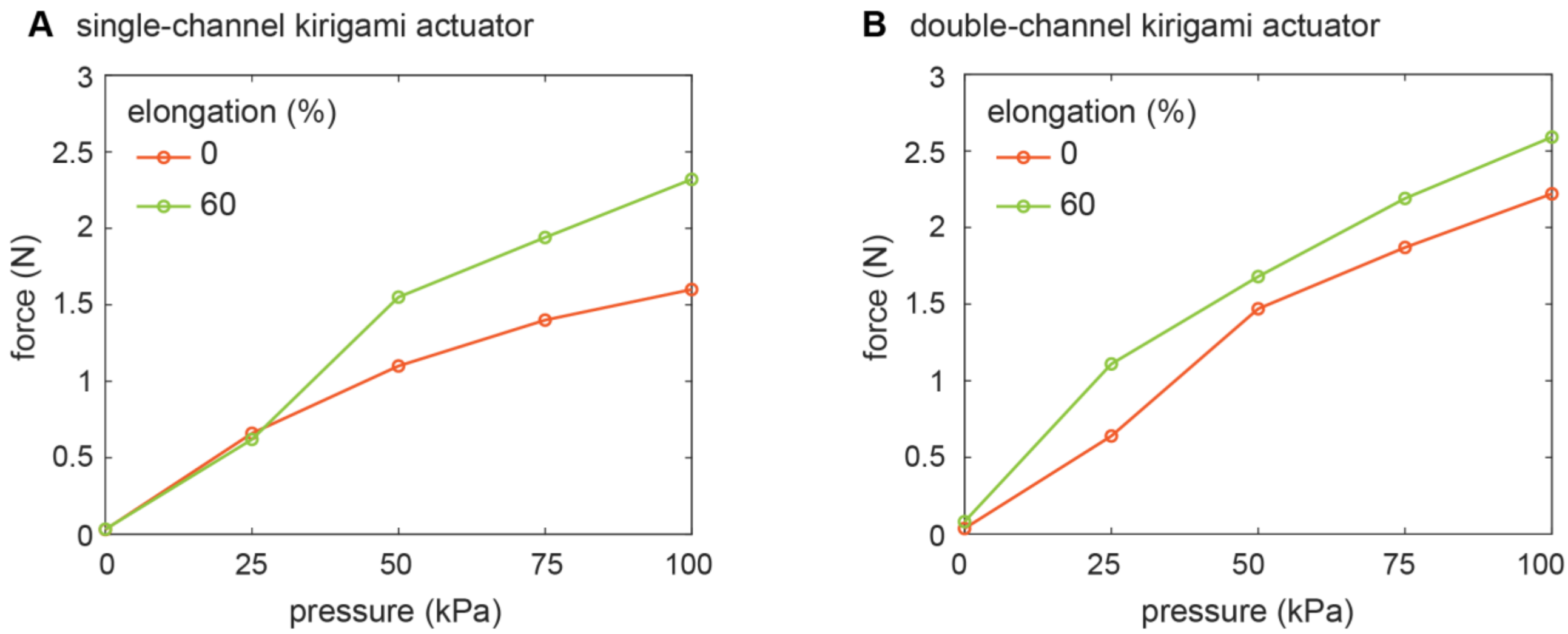

**Figure S5. Blocked force results**. (A) Single-channel and (B) double-channel kirigami actuators were tested under blocked conditions (0% and 60% elongation) across incrementally increased pressure levels. The force values represent the average measurements taken over a 10-second period at each pressure step, sampled at 0.01-second intervals.

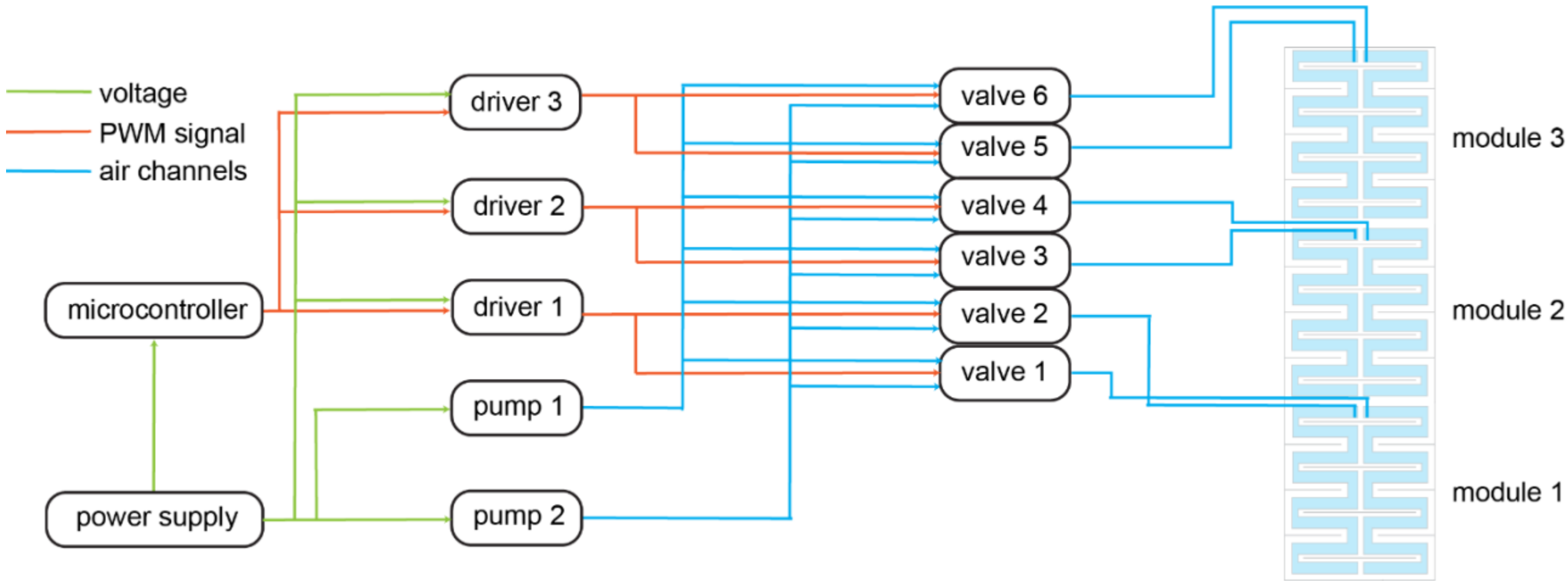

**Figure S6. The schematics of pneumatic and electronic connections**. The microcontroller, drivers, and pumps were powered separately. Each valve is responsible for controlling the inflation and deflation cycle of an individual air channel. Green lines show the power connections, orange lines show pulse-width modulation (PWM) signals sent through microcontroller to valves through drivers, and blue lines display the air flow.

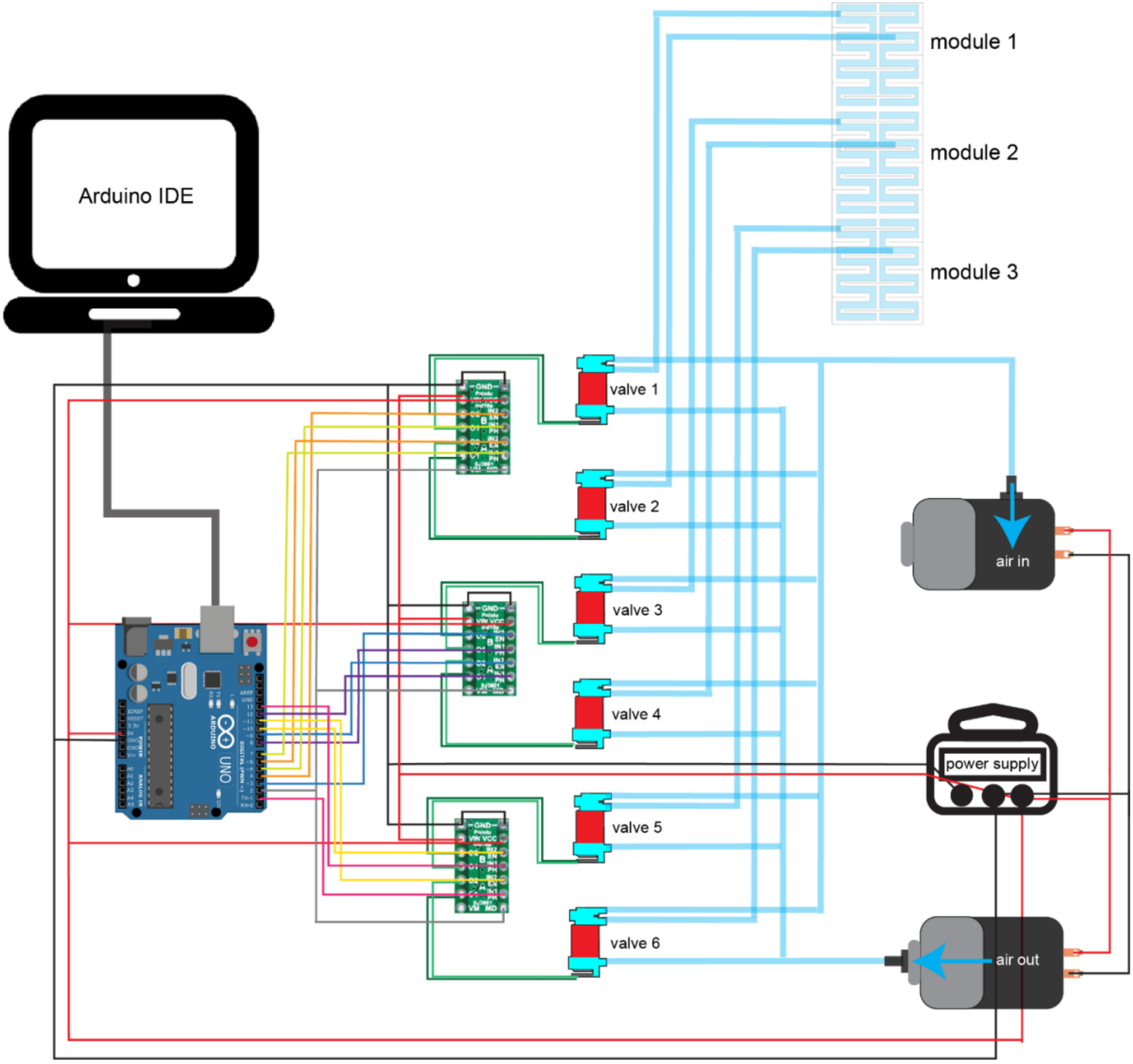

**Figure S7. Control architecture**. The wires indicate the connections of the microcontroller, drivers, and the 3-way valves (colored thin lines). The blue thick lines show the pneumatic connection between the pumps, valves and the air channels.

**Note S3. Friction analysis**

We performed friction tests on both single-channel and double-channel kirigami actuators by measuring the friction force in both forward and backward directions, achieved by simply changing their orientation. The experiments were conducted in two states: the initial (unactuated) state and the inflated state, pressurized to 100 kPa using a microfluidic pressure controller system (Fluigent LineUp™).

During actuator sliding, the friction behavior exhibited distinct stick-slip characteristics. To analyze this, we applied a previously developed post-processing method for stick-slip friction [2]. Static friction was estimated by averaging the dominant peaks, while kinetic friction was determined by averaging the lowest troughs of the friction force within a 10-second sliding window. MATLAB was used to identify the ten most prominent peaks for each condition. The obtained friction coefficients were reported based on the maximum values.

**Note S4. Description of supporting videos**

Video S1. Comparison of the deformation between single-channel kirigami actuator and air-pouch without cuts.

Video S2. Contraction and bending behavior of double-channel kirigami actuator.

Video S3. Deformation of three-module kirigami crawler under various actuation protocols.

Video S4. Comparison of experimental and simulated behaviors of single-channel kirigami actuator contraction; double-channel kirigami actuator contraction; and double-channel kirigami actuator bending.

Video S5. Linear locomotion of single module single-channel and double-channel kirigami actuators on coarse foam.

Video S6. Turning of double-channel kirigami actuator on coarse foam.

Video S7. Serpentine and rectilinear locomotion of two module and three module kirigami crawlers under different actuation phase shift configurations on coarse foam.

Video S8. Rectilinear and turning gaits of three-module kirigami crawler, completing an S-shaped path on coarse foam.

Video S9. Three-module kirigami robot traverses an inclined coarse foam surface and passes through a narrow channel. Side views of three-module kirigami crawler with and without actuation phase shifting.

Video S10. Rectilinear locomotion of three-module kirigami crawler on asphalt.

Video S11. Rectilinear locomotion of three-module kirigami crawler on inclined concrete pavement.

Video S12. Rectilinear locomotion of three-module kirigami crawler on concrete pavement.

Video S13. Rectilinear locomotion of three-module kirigami crawler on metal grating.

Video S14. Rectilinear locomotion of three-module kirigami crawler on curved surfaces.